\documentclass{article}
\usepackage{spconf,amsmath,graphicx,hyperref}

\usepackage{xcolor}
\usepackage{booktabs}
\usepackage{multirow}
\usepackage{siunitx}
\usepackage{makecell}
\usepackage{array,verbatim}
\usepackage[normalem]{ulem}

\title{Double-stream registration with pyramid fusion for HDR video with alternating exposures}
\name{Onofre Martorell \qquad Ivan Pereira-Sánchez \qquad Antoni Fuentes \qquad Antoni Buades \thanks{ This work is part of the MoMaLIP project PID2021-125711OB-I00 funded by MICIU/AEI/10.13039/501100011033 and the European Union NextGeneration EU/PRTR. The authors gratefully acknowledge the computer resources at Artemisa, funded by the EU ERDF and Comunitat Valenciana and the technical support provided by IFIC (CSIC-UV).}}
\address{%
Dept. of Mathematics and Computer Science, \\
Universitat de les Illes Balears, Cra. de Valldemossa km 7.5, Palma, 07122, Illes Balears, Spain\\
Institute of Applied Computing and Community Code (IAC3),\\
Universitat de les Illes Balears, C/ Blaise Pascal 7, Parc BIT, Palma, 07121, Illes
Balears, Spain
}

\begin{document}
%
\maketitle
\begin{abstract}
High dynamic range (HDR) video reconstruction from al\-ter\-na\-ting-exposure sequences remains challenging, especially in regions with extreme luminance variation. We propose a novel HDR reconstruction framework based on dual-stream registration and accurate pyramid fusion. Given three consecutive frames, our method computes optical flow directly with the central frame, while introducing a complementary midpoint displacement strategy to handle cases with severe overexposition. A pyramid fusion stage then merges the resulting radiance and LDR images into a final HDR output. Experimental results demonstrate that our approach consistently outperforms state-of-the-art methods. 
\end{abstract}


\section{Introduction}
The demand for high-quality video content with high resolution and high dynamic range (HDR) has grown rapidly in the last years. However, standard consumer cameras and smartphones capture only a narrow low dynamic range (LDR), resulting in severe detail loss,  saturation in overexposed regions and noise in underexposed areas. While specialized hardware solutions \cite{tocci2011versatile, mcguire2007optical_splitting}  (e.g., multi-sensor designs or beam splitters) can directly capture HDR video, they remain  expensive and difficult to include in daily use devices.

To overcome these constraints, a highly practical and cost-effective computational alternative is reconstructing HDR videos from LDR sequences captured with alternating exposures. This approach typically records consecutive LDR frames with alternating exposure times (usually two or three) to later process the video and recover the full dynamic range at each frame. However, fusing alternately exposed frames remains a highly ill-posed challenge.

Most image HDR imaging pipelines \cite{liu2022ghost} are designed to use a reference frame with medium-exposure to which the other exposures are geometrically registered.  These methods cannot be directly extended, since  each frame needs to be used as reference even if it is significantly over or under-exposed. Such an application leads to 
severe temporal inconsistency and flickering across the reconstructed video. HDR video methods must therefore be explicitly designed.




Deep learning has driven rapid progress in HDR video, with methods evolving from feature-space attention~\cite{chung2023lan} and HDR-supervised optical flow~\cite{xu2024hdrflow}, to exposure-completion via interpolation~\cite{cui2024exposure}, unified exposure representations avoiding explicit alignment~\cite{liao2026lrhdr}, and physically-guided cross-exposure motion fields~\cite{yue2026f2hdr}, underscoring how substantially neural approaches have reshaped this problem in just a few years.

\medskip
{
We propose a new model based on {dual-stream} registration and accurate pyramid fusion. The  method takes three consecutive frames from an alternating-exposure video as input.
For areas that are well exposed in consecutive frames, direct optical flow computed on the equalized LDR frames enables accurate registration. However, the main source of error arises when an area is over-exposed in one of the frames.  For this reason, we complement the registration between each neighboring frame and the central frame using the midpoint of the displacement vector obtained by applying optical flow directly between the two neighboring frames. This latter displacement enables accurate registration, especially when the central frame has a long exposure.
This method relies on the temporal coherence of the flow, assuming that it is approximately linear for each pixel across consecutive frames.
A pyramid fusion of the five radiance and LDR images, that is, central and two neighboring frames with each of the two registered ones, achieves the desired HDR.
The experimental section illustrates how dual-stream registration allows us to take the best of each registration strategy, achieving significantly better performance than the current state of the art.
}

\medskip

The remainder of this paper is structured as follows. Section \ref{sec:related_work} reviews the existing literature on HDR video reconstruction. Section \ref{sec:method} details our proposed methodology, followed by Section \ref{sec:results}, which presents a comparative evaluation against state-of-the-art methods. Finally, Section \ref{sec:conclusions} concludes the paper and outlines directions for future research.

\section{Related Work}
\label{sec:related_work}

There is an extensive literature for HDR imaging, we refer the reader to the comprehensive reviews in both classical \cite{tursun2015state} and deep learning methods~\cite{martins2025review}.  In the rest of the section, we focus on video HDR methods.  

\subsection{Classical methods for  HDR Video Reconstruction}

Kang et al.~\cite{kang2003high} proposed the first HDR video reconstruction method, aligning neighbouring frames to the reference frame through a combination of global transformation and optical flow, then merging the aligned frames via a weighted average that accounts for exposure and optical flow errors. Most subsequent methods build on this general framework, refining either the registration or the merging strategy.

Among classical approaches, Mangiat et al.~\cite{mangiat2010high}, Li et al.~\cite{li2016maximum}, and Van Vool et al.~\cite{van2020high} focus on improving image registration accuracy across frames of differing exposure times, yielding better reconstruction quality. Other works, including Kalantari et al.~\cite{kalantari2013patch}, Gryaditskaya et al.~\cite{gryaditskaya2015motion}, and Buades et al.~\cite{buades2023hdr}, instead formulate the problem within an optimization framework, obtaining the reconstruction as the solution to an energy minimization.

\subsection{Deep learning methods for  HDR Video Reconstruction}

Early deep-learning approaches relied on multi-scale CNNs for optical flow  and frame blending~\cite{kalantari2019deep}, later refined with deformable convolutions~\cite{chen2021hdr}. To handle cross-exposure alignment, LAN-HDR~\cite{chung2023lan} leverages luminance-based structural attention and gated convolutions for saturated region recovery, while HDRFlow~\cite{xu2024hdrflow} trains an efficient large-kernel flow network under an HDR-domain alignment loss. NECHDR~\cite{cui2024exposure} takes a different route, interpolating missing LDR frames to ensure a complete exposure set before fusion. More recently, LRHDR~\cite{liao2026lrhdr} bypasses explicit alignment by mapping features into an exposure-independent domain with adaptive sparse fusion, and F2HDR~\cite{yue2026f2hdr} refines generic optical flow using a motion-aware network to suppress ghosting artifacts.



\section{Method}
\label{sec:method}

The proposed method takes as input a multi-exposure video sequence  $\{L_t \ \mid t = 1, \dots, N\}$ with two different exposures. 
In order to compute the HDR of a certain frame $L_t$ we will use its previous and posterior frames, $L_{t-1}$ and $L_{t-1}$ respectively.
The current approach can be modified to take into account a larger number of different exposures and frames.

Let $H_i,\ i\in \{t-1, t, t+1\}$ denote the corresponding radiance frames. The radiance is either computed by inverting the CRF using Devebec and Malik \cite{debevec2008recovering} or by using a gamma function to obtain a linear radiance \cite{kalantari2019deep, wu2018deep}. In both cases, dividing the obtained images after the CRF by the corresponding exposure time provides a set of images that can be compared and combined.

\medskip

For a target central frame $L_t$, we register its two temporal neighbors ($L_{t-1}$ and $L_{t+1}$) using two distinct alignment strategies to capture complementary motion cues:

\medskip

{
\begin{itemize}
    \item \textbf{Direct Optical Flow ($W^{t \pm 1 \to t}$):} The neighboring frames are warped using the  flow computed directly between each respective neighbor and  the central. 
    \item \textbf{Midpoint Motion Vectors ($\frac12 W^{t \pm 1 \to t \mp 1}$):} The neighboring frames are warped using half of the motion vector computed exclusively between the two outer frames.
\end{itemize}

Consequently, the network input aggregates five components: the central frame $(L_t,H_t)$, alongside the directly warped neighbors $(L_{t-1}^{W^{t - 1 \to t}},H_{t-1}^{W^{t - 1 \to t}})$, $(L_{t+1}^{W^{t + 1 \to t}},H_{t+1}^{W^{t + 1 \to t}})$ and midpoint-warped neighbors  $(L_{t-1}^{\frac12 W^{t - 1 \to t+1}},H_{t-1}^{\frac12 W^{t - 1 \to t+1}})$ and $(L_{t+1}^{\frac12 W^{t + 1 \to t-1}},H_{t+1}^{ \frac12 W^{t + 1 \to t-1}})$.

\medskip

We adopt a multi-scale pyramid approach to fuse the registered radiances and LDR images in order to obtain the desired output radiance $S_t$. Pyramid details at each scale are combined using a U-Net architecture.

\medskip
\medskip
{\noindent \bf Flow  Registration.} In order to compute the flow between images $L_i$ and
$L_j$, we first photometrically calibrate the color values of the darker
 image to look alike the brighter one. The flow is then computed with the optical
 flow algorithm PWC-Net  \cite{Sun2018PWC-Net}. We adopt a forward warping strategy using \cite{niklaus2020softmax}. Both the LDR and radiance images are transformed.

\medskip
\medskip
{\noindent \bf Pyramid Fusion.} We adopt a classical Laplacian pyramid approach in which both the LDR and radiance image are decomposed using a  feature pyramid extractor}.  We use a GridNet \cite{fourure2017gridnet} for the fusion at each scale, which is a generalization of a Unet.



\medskip
\medskip
{\noindent \bf Loss. }Following previous works, \cite{chung2023lan,xu2024hdrflow, yue2026f2hdr,  chen2021hdr} we use the differentiable $\mu$-law function as tone-mapping 

\begin{equation}
\label{tonemap_mu}
    \mathcal{T}(H) =  \frac{\log(1+\mu H)}{\log(1+H)},
\end{equation}
and we compute the reconstruction loss with an $L_1$ loss
\begin{equation}
    \mathcal{L} = \|\mathcal{T}(H_{final}) - \mathcal{T}(H_{gt})\|_1.
\end{equation}

\section{Results}
\label{sec:results}

\begin{figure*}[th]
\centering

\setlength{\tabcolsep}{2pt}
\begin{tabular}{p{8pt} c c c c c}
\rotatebox[origin=lt]{90}{\small Input} && \includegraphics[width=3.4cm]{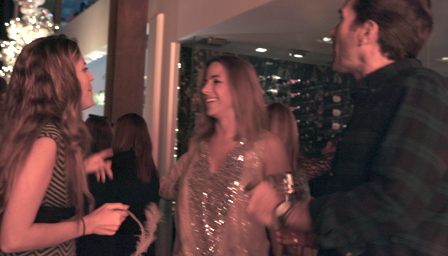} &
\includegraphics[width=3.4cm]{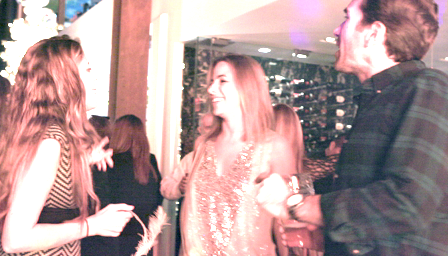} &
\includegraphics[width=3.4cm]{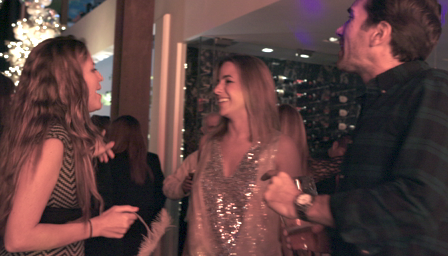} & \\
\rotatebox[origin=lt]{90}{\small Result} & \includegraphics[width=3.4cm]{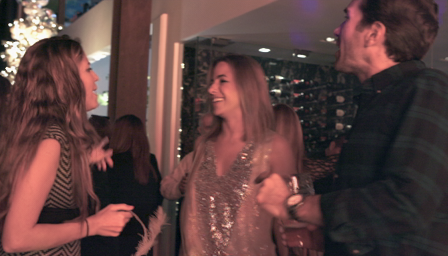} &
\includegraphics[width=3.4cm]{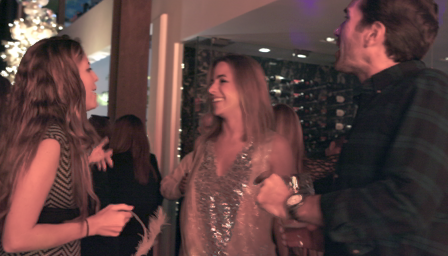} &
\includegraphics[width=3.4cm]{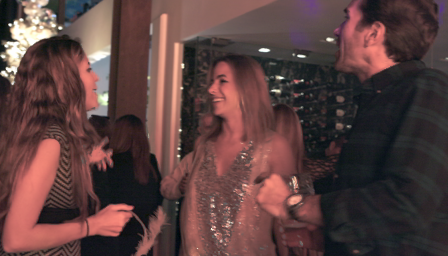} &
\includegraphics[width=3.4cm]{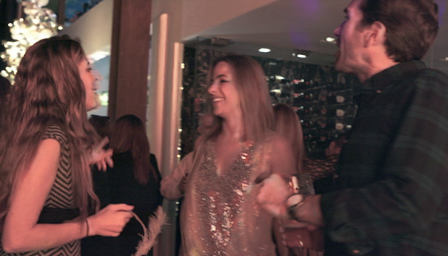} &
\includegraphics[width=3.4cm]{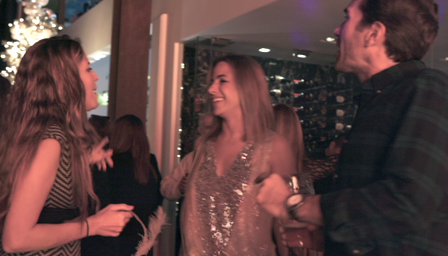}\\

\rotatebox[origin=lt]{90}{\small Error} & 
\includegraphics[width=3.4cm]{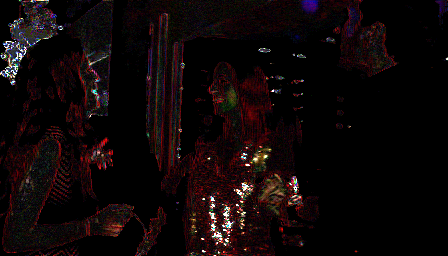}&
\includegraphics[width=3.4cm]{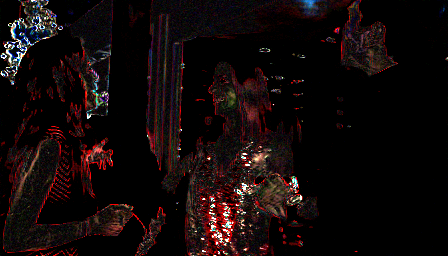}&
\includegraphics[width=3.4cm]{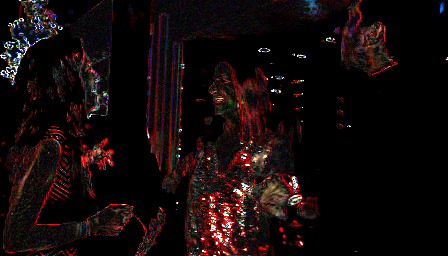}&
\includegraphics[width=3.4cm]{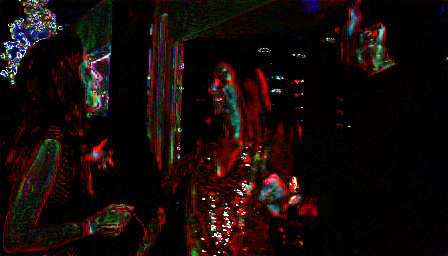}&
\includegraphics[width=3.4cm]{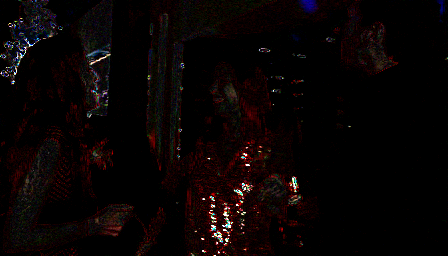}\\

\rotatebox[origin=lt]{90}{\small Detail} & 
\includegraphics[width=3cm]{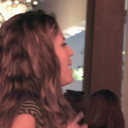}&
\includegraphics[width=3cm]{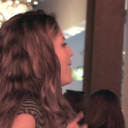}&
\includegraphics[width=3cm]{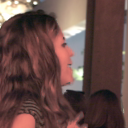}&
\includegraphics[width=3cm]{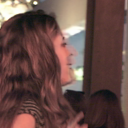}&
\includegraphics[width=3cm]{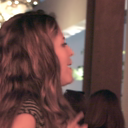}\\
& Deep HDR \cite{chen2021hdr}& F2HDR \cite{yue2026f2hdr} & HDR Flow  \cite{xu2024hdrflow} & LAN HDR \cite{chung2023lan} & Ours\\
\end{tabular}

\caption{Comparison on Vimeo dataset centered on a long exposure frame. The error between each output and the ground truth illustrates the superiority of the proposed method. This is also noticleable in the women's face in the detail images.}\label{fig:1}
\end{figure*}

\begin{figure*}[th]
\centering

 \begin{tabular}{p{8pt} c c c c c} \setlength{\tabcolsep}{2pt}
\rotatebox[origin=lt]{90}{\small Input} &&
\includegraphics[trim=0 0 500pt 0,clip, width=3cm] {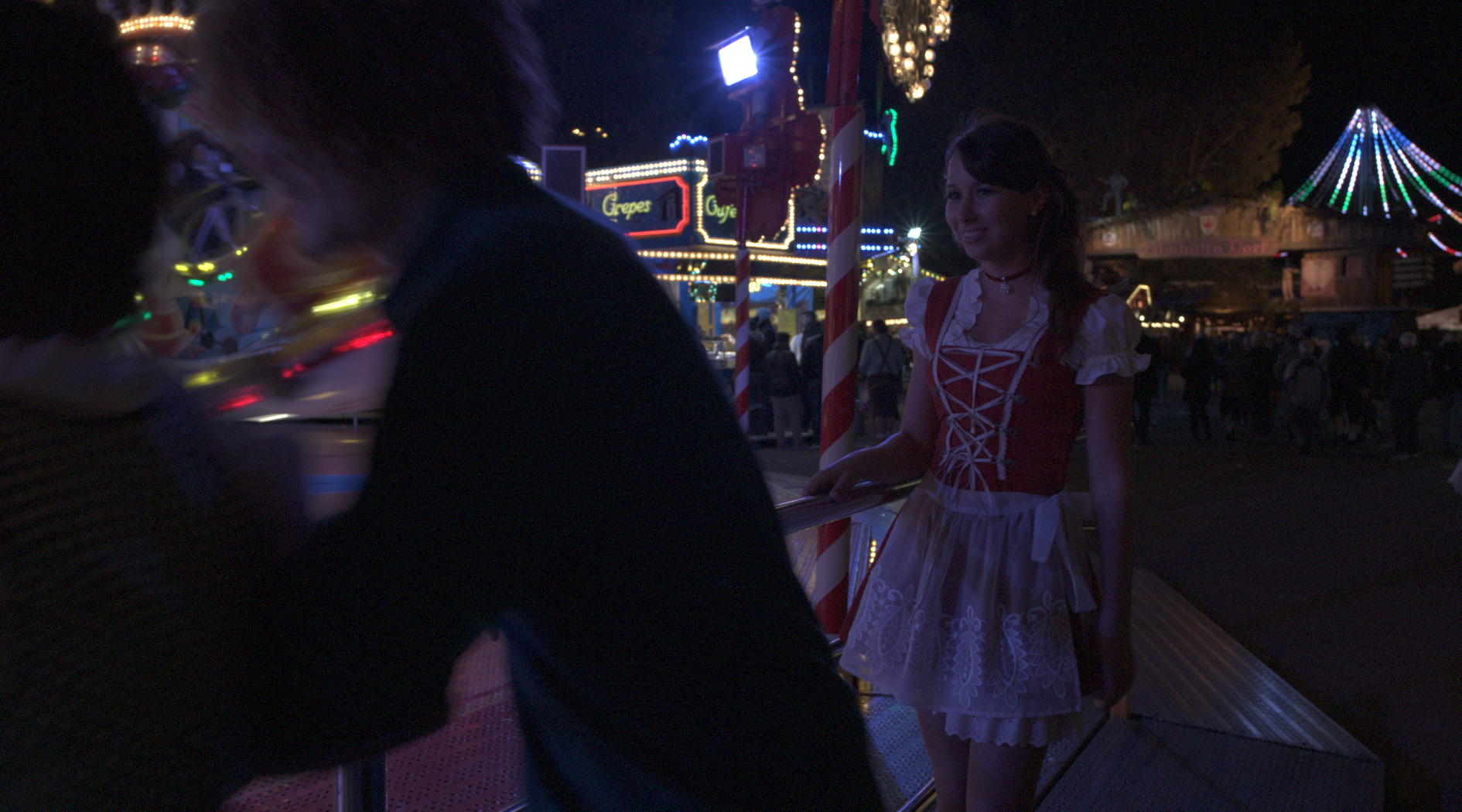} &
\includegraphics[trim=0 0 500pt 0,clip, width=3cm]{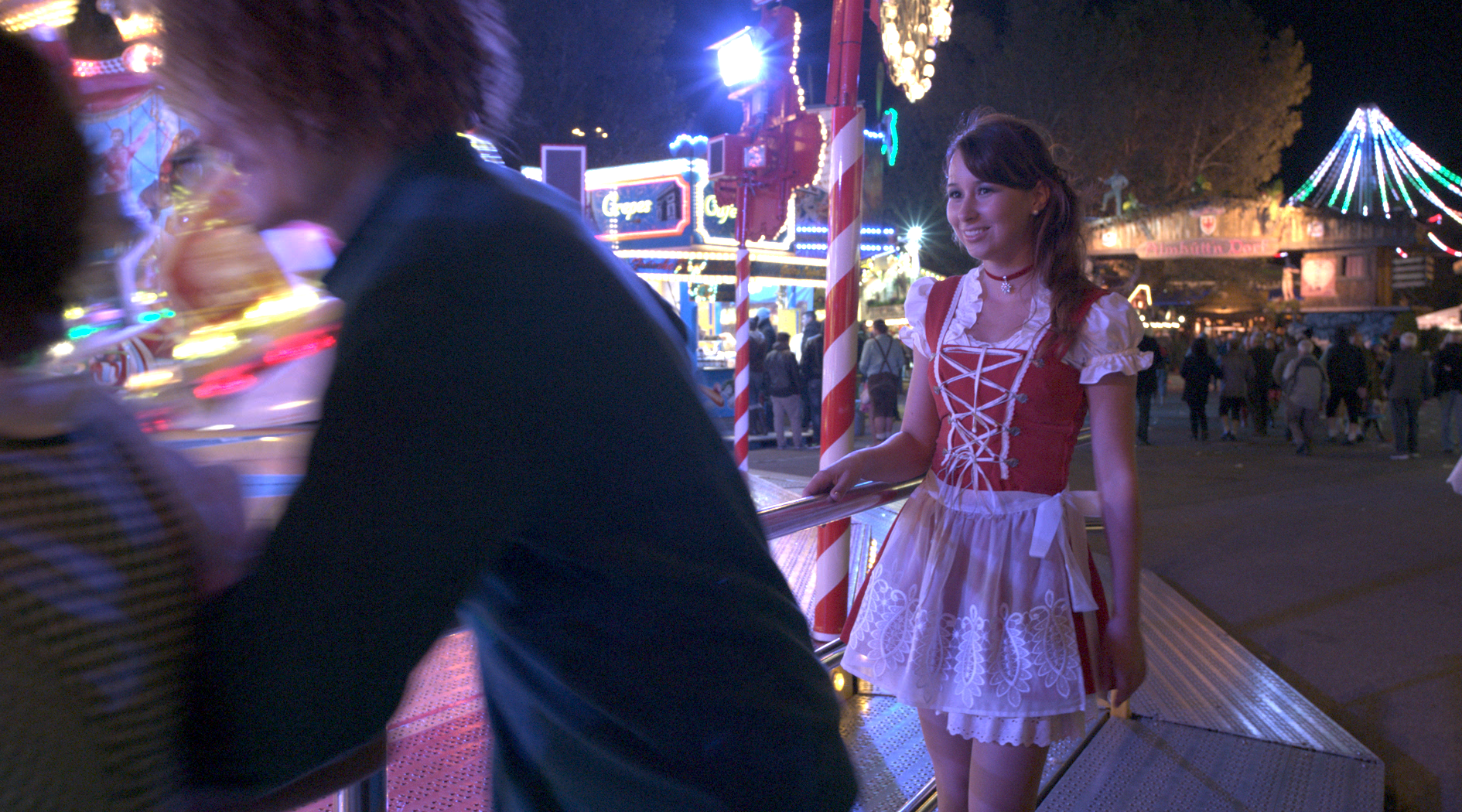} &
\includegraphics[trim=0 0 500pt 0,clip, width=3cm]{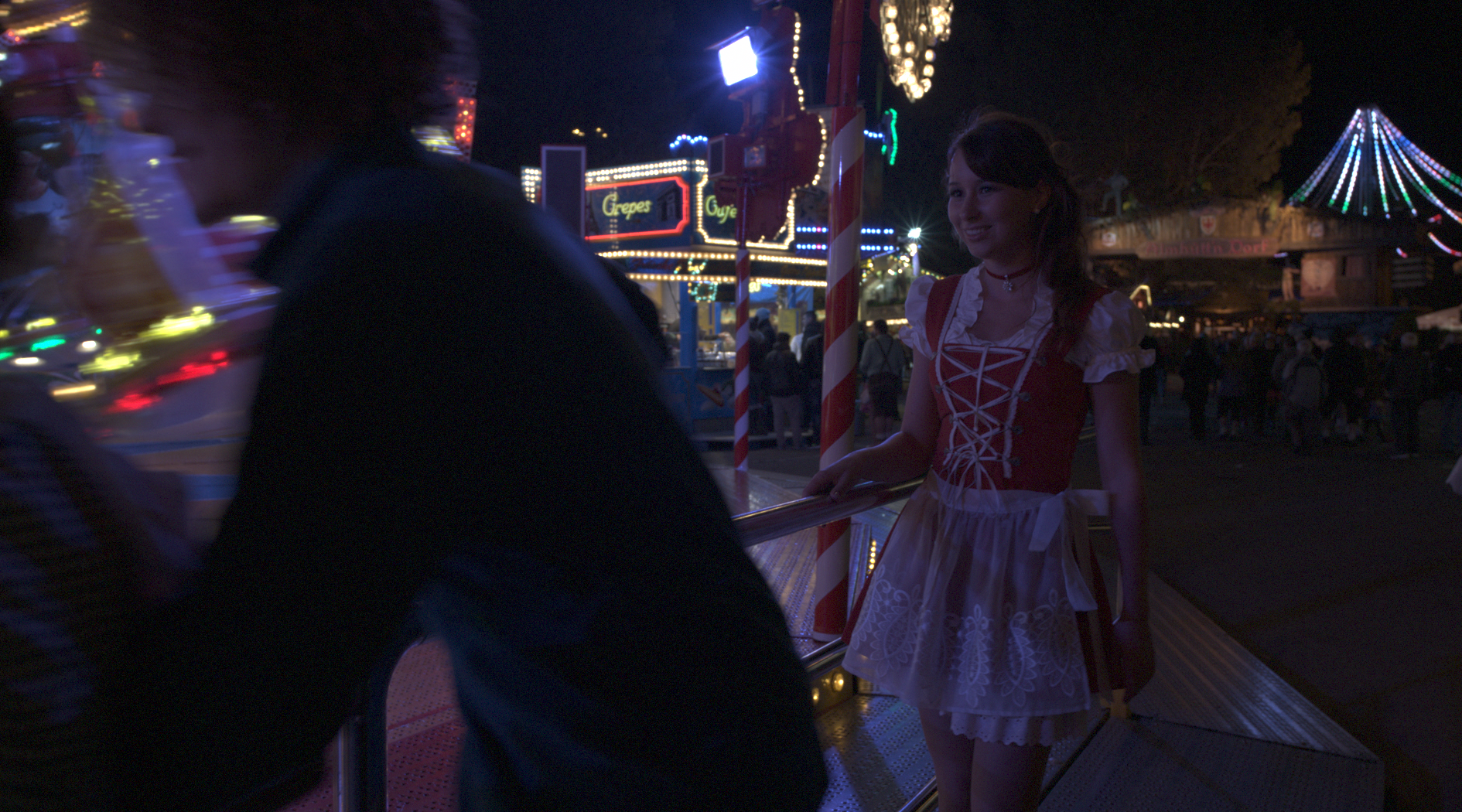} & \\

\setlength{\tabcolsep}{2pt}
\rotatebox[origin=lt]{90}{\small Result} &
\includegraphics[trim=0 400 1000pt 0,clip, width=3cm]{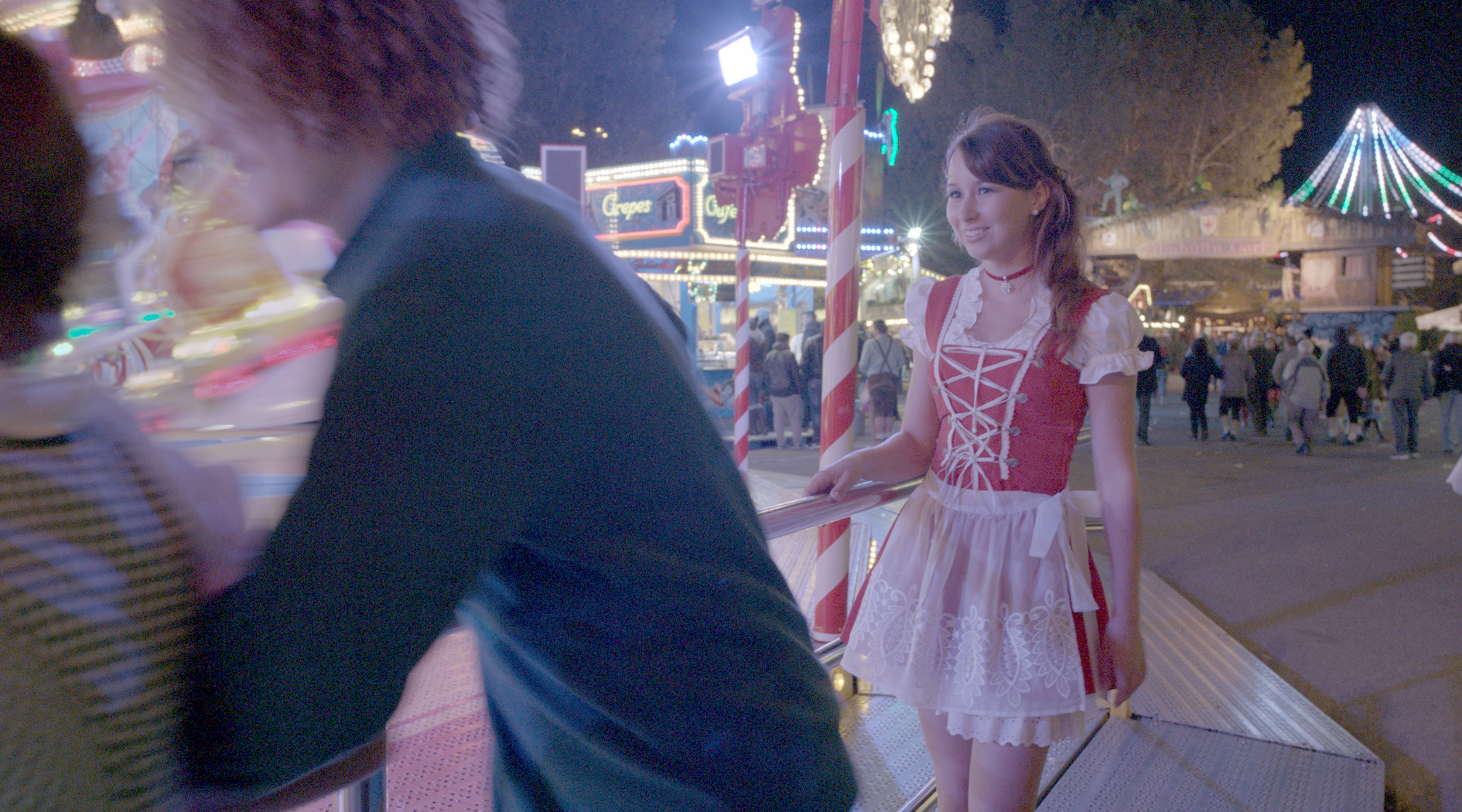} &
\includegraphics[trim=0 400 1000pt 0,clip, width=3cm]{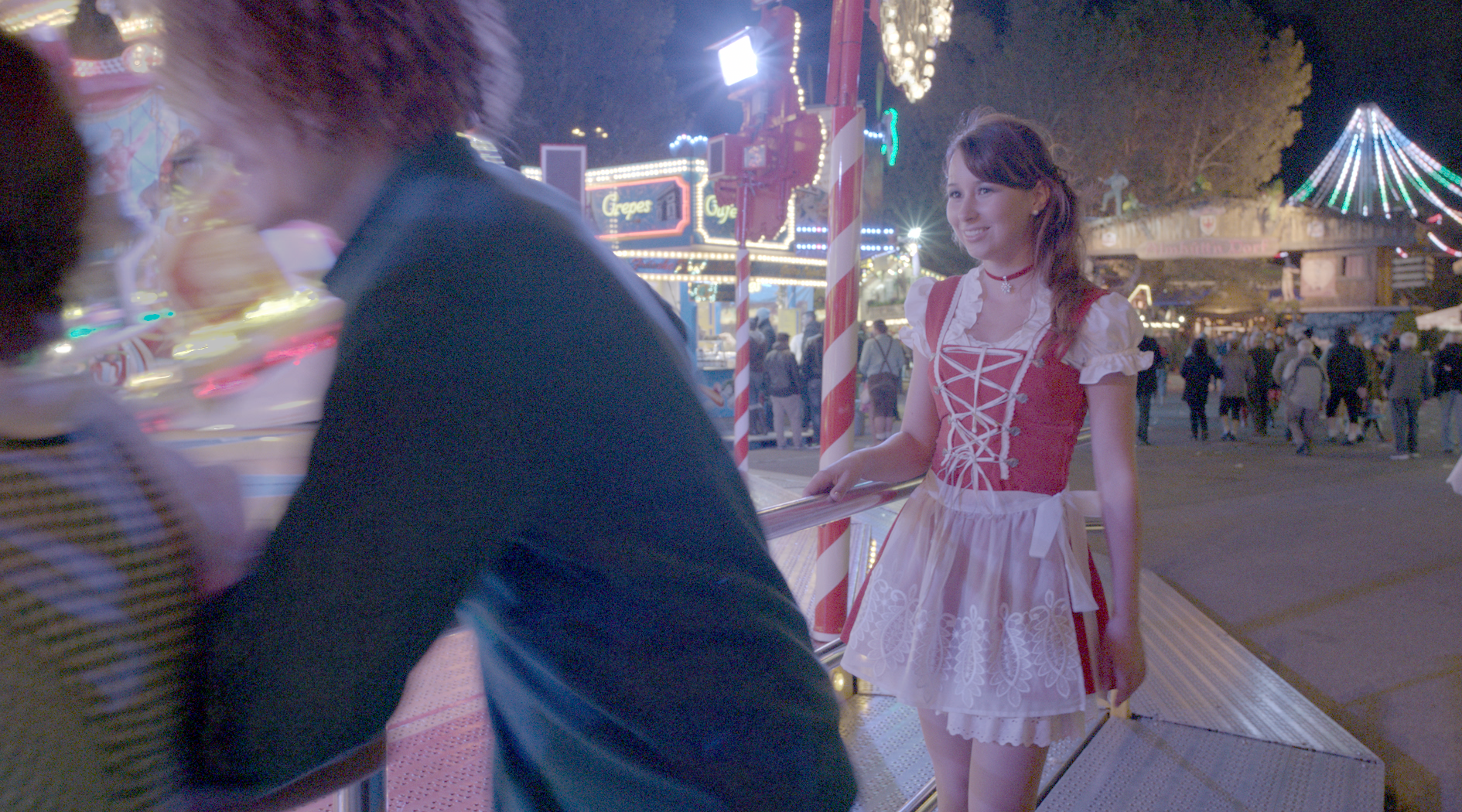}&
\includegraphics[trim=0 400 1000pt 0,clip, width=3cm]{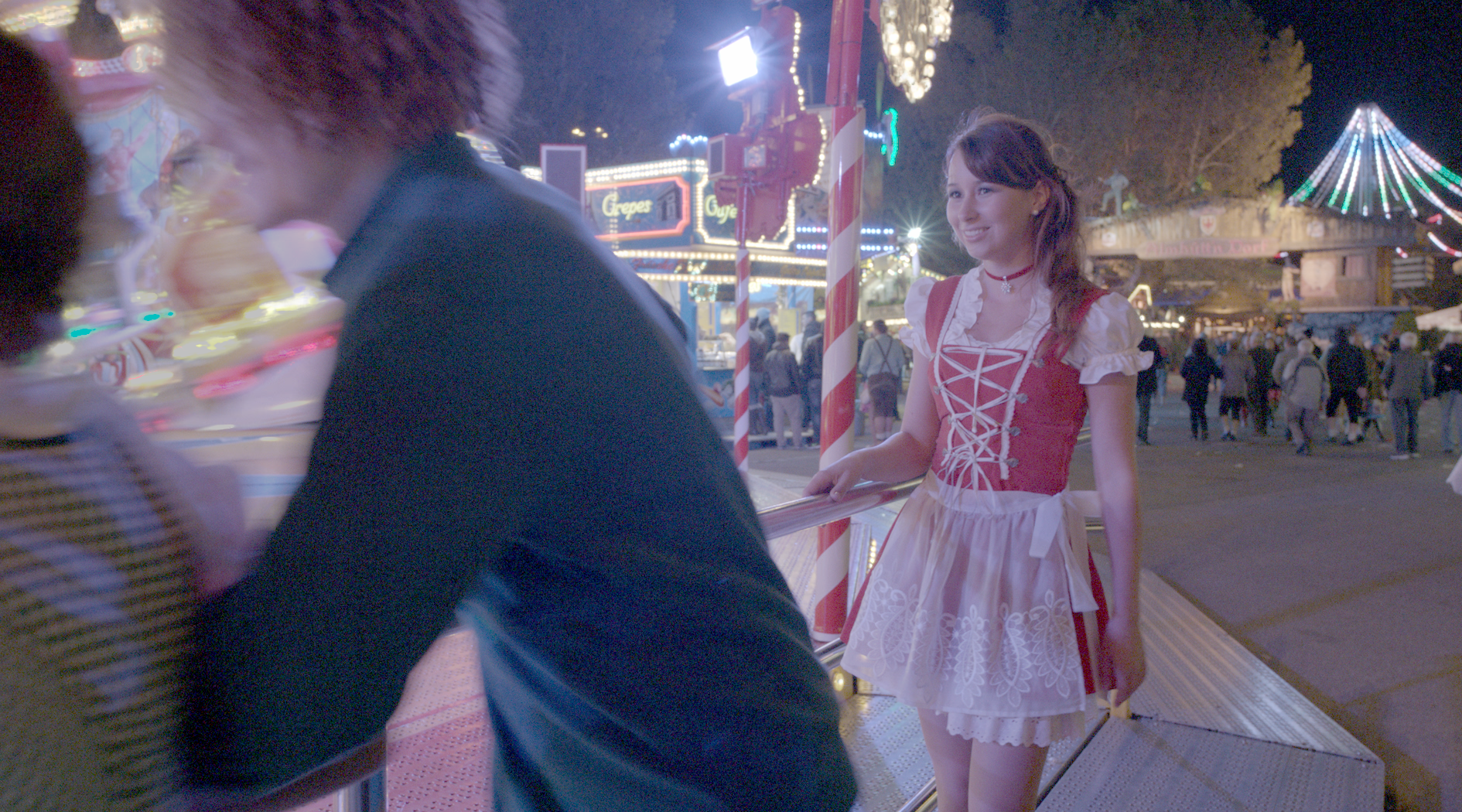}&
\includegraphics[trim=0 400 1000pt 0,clip, width=3cm]{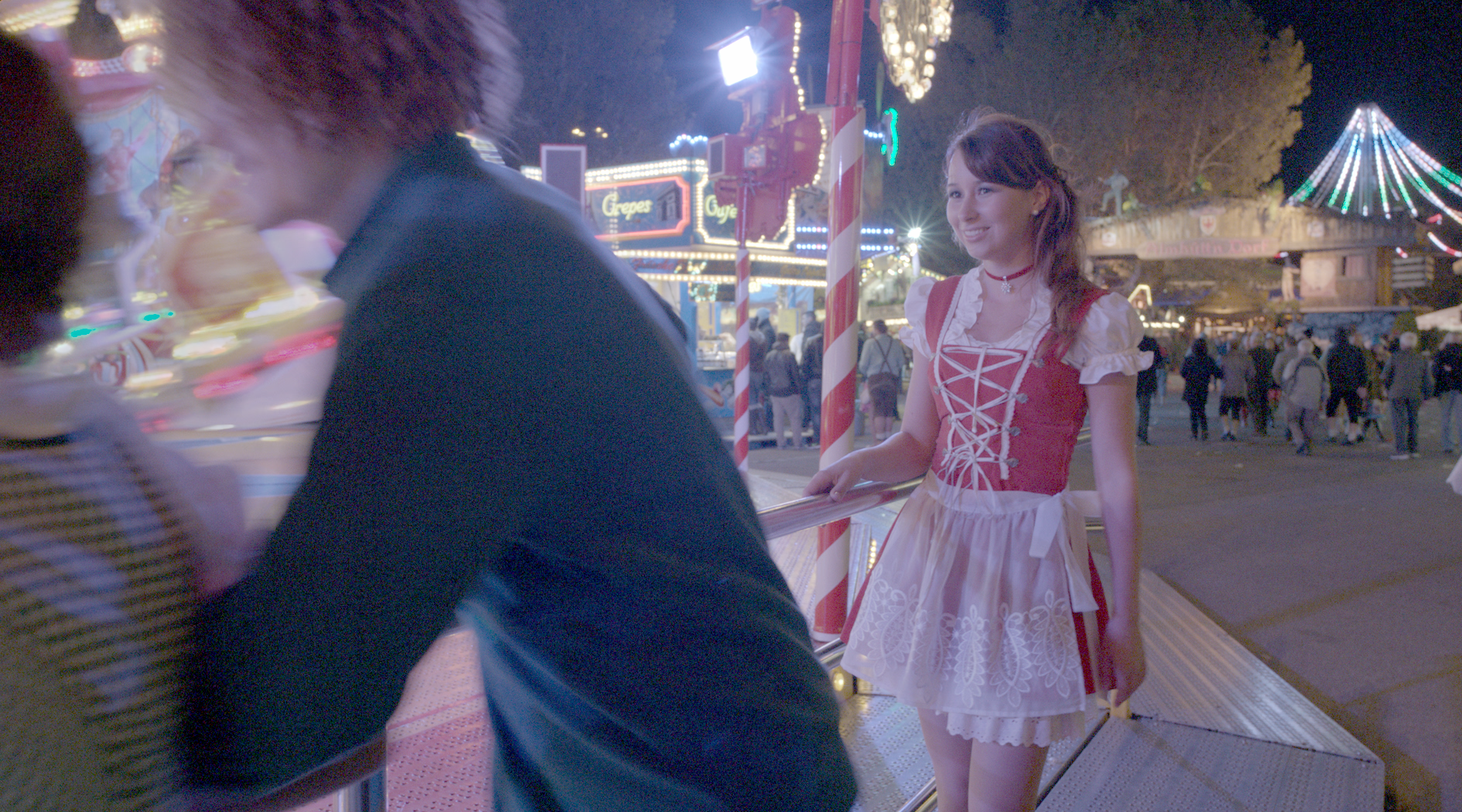}&
\includegraphics[trim=0 400 1000pt 0,clip, width=3cm]{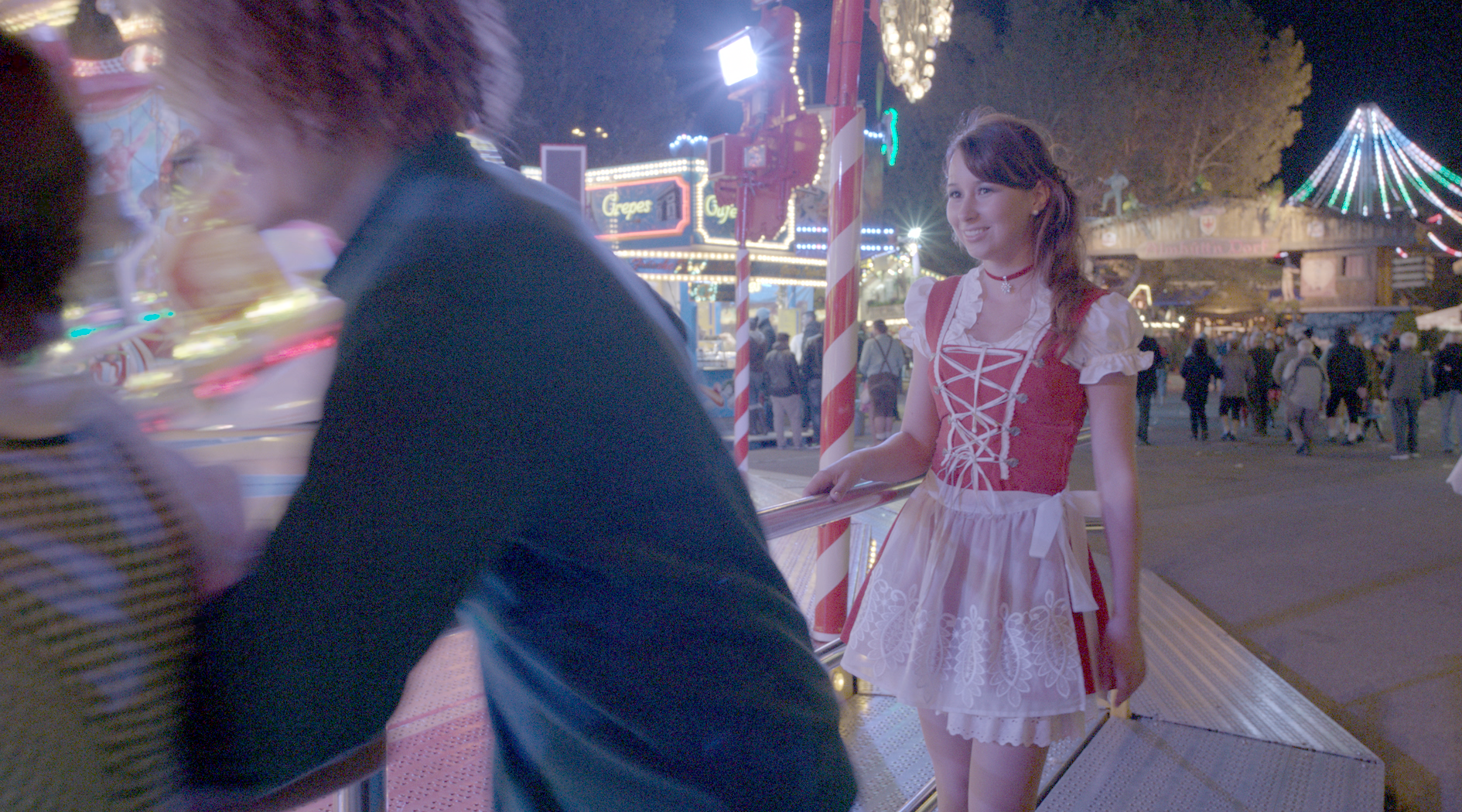} \\

\rotatebox[origin=lt]{90}{\small Error} &
\includegraphics[trim=0 400 1000pt 0,clip, width=3cm]{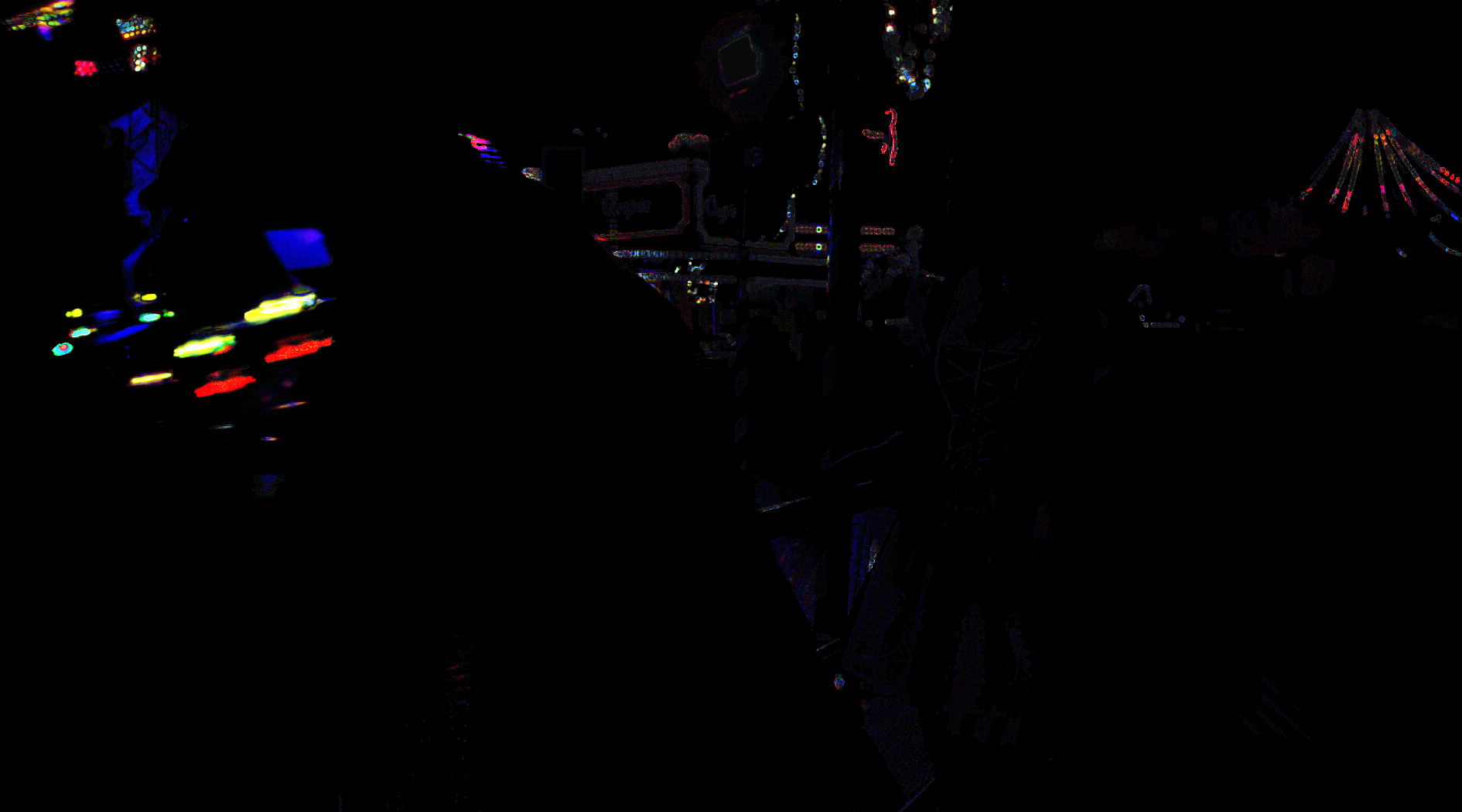} &
\includegraphics[trim=0 400 1000pt 0,clip, width=3cm]{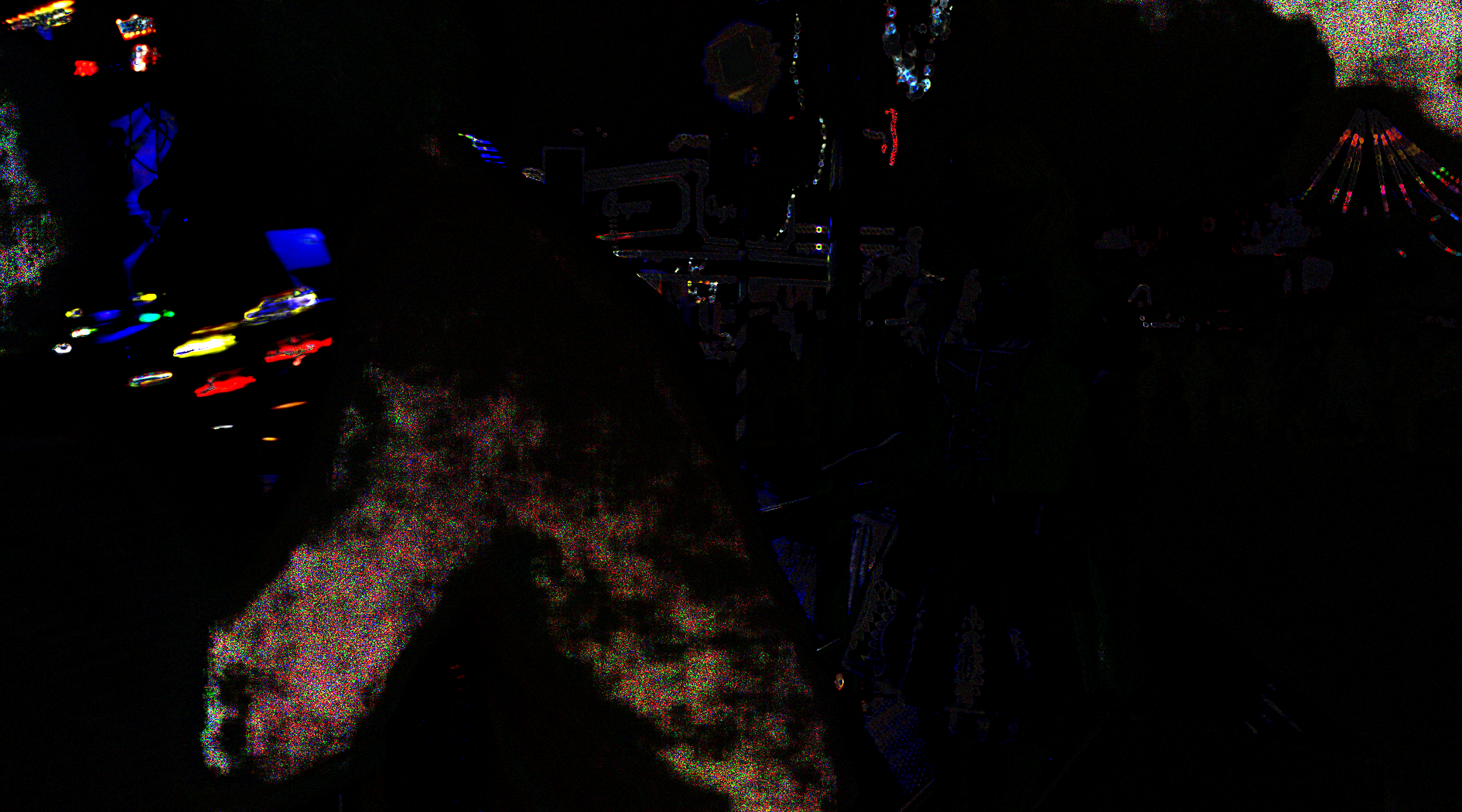}&
\includegraphics[trim=0 400 1000pt 0,clip, width=3cm]{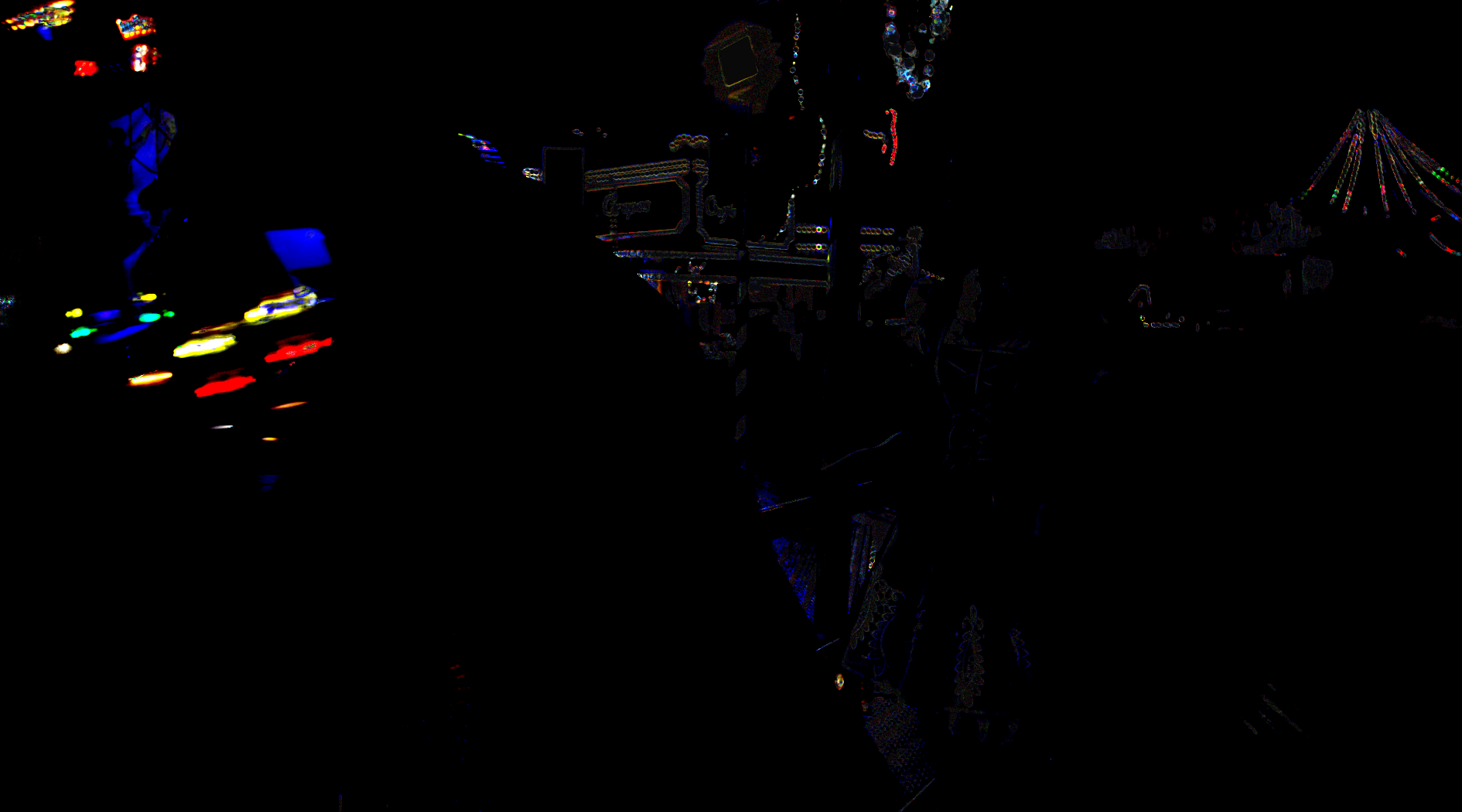}&
\includegraphics[trim=0 400 1000pt 0,clip, width=3cm]{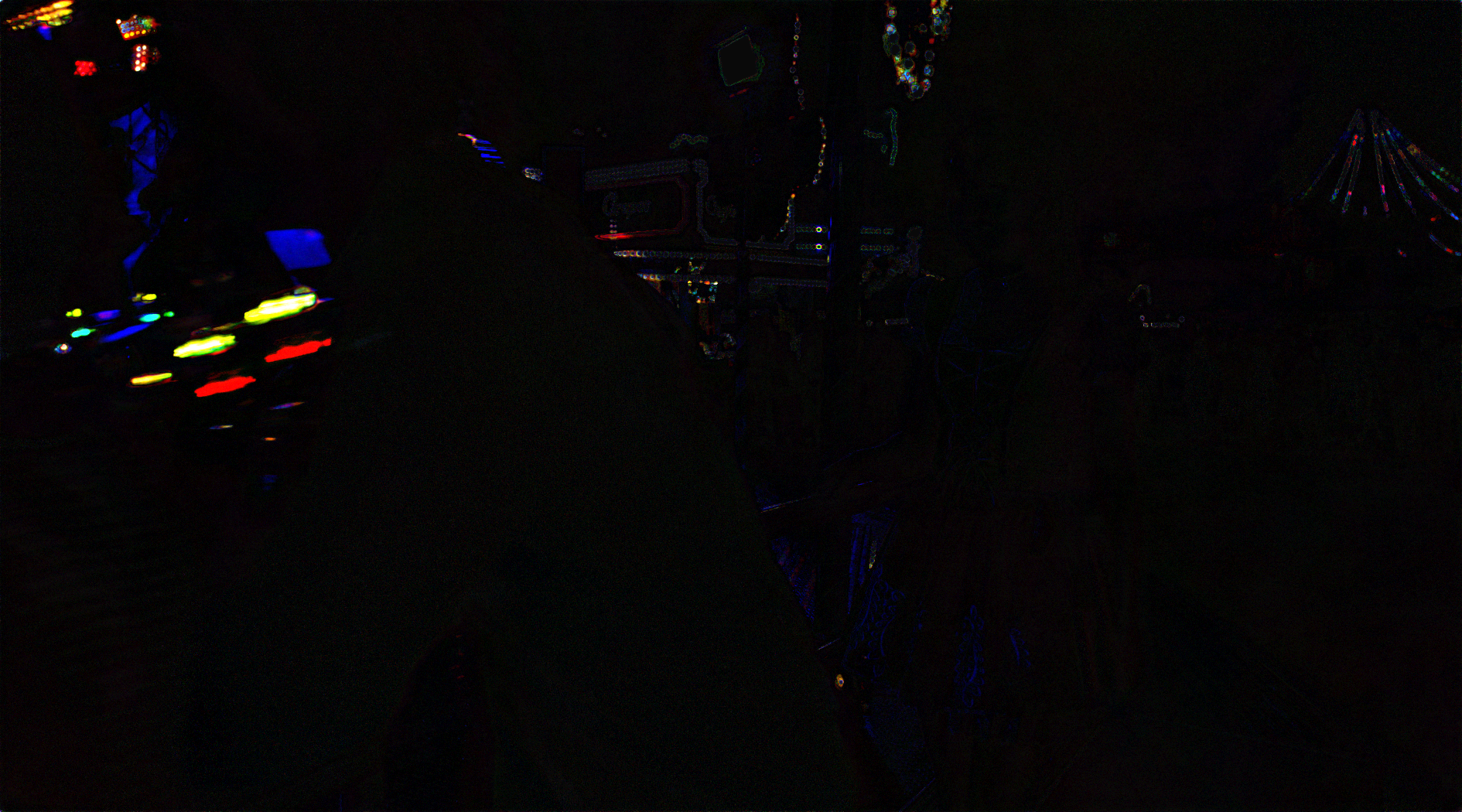}&
\includegraphics[trim=0 400 1000pt 0,clip, width=3cm]{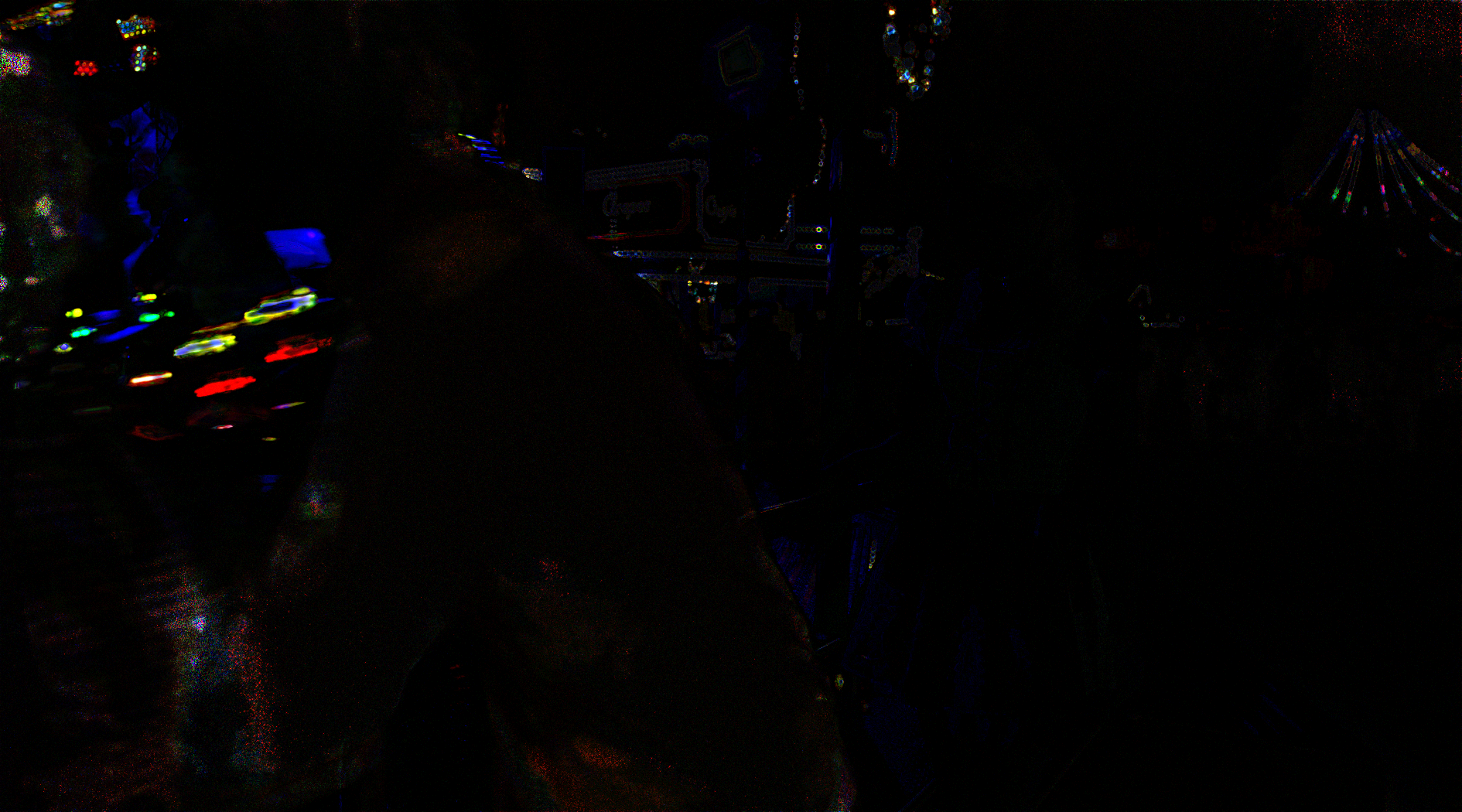}\\
& Chen et al. \cite{chen2021hdr}& F2HDR \cite{yue2026f2hdr} & HDR Flow  \cite{xu2024hdrflow} & LAN HDR \cite{chung2023lan} & Ours\\
\end{tabular}

\caption{Comparison on Stuttgart dataset centered on a long exposure frame. }\label{fig:2}
\end{figure*}

\begin{figure}[th]
\centering

\includegraphics[trim=600pt 100 400 100,clip, width=2.8cm]{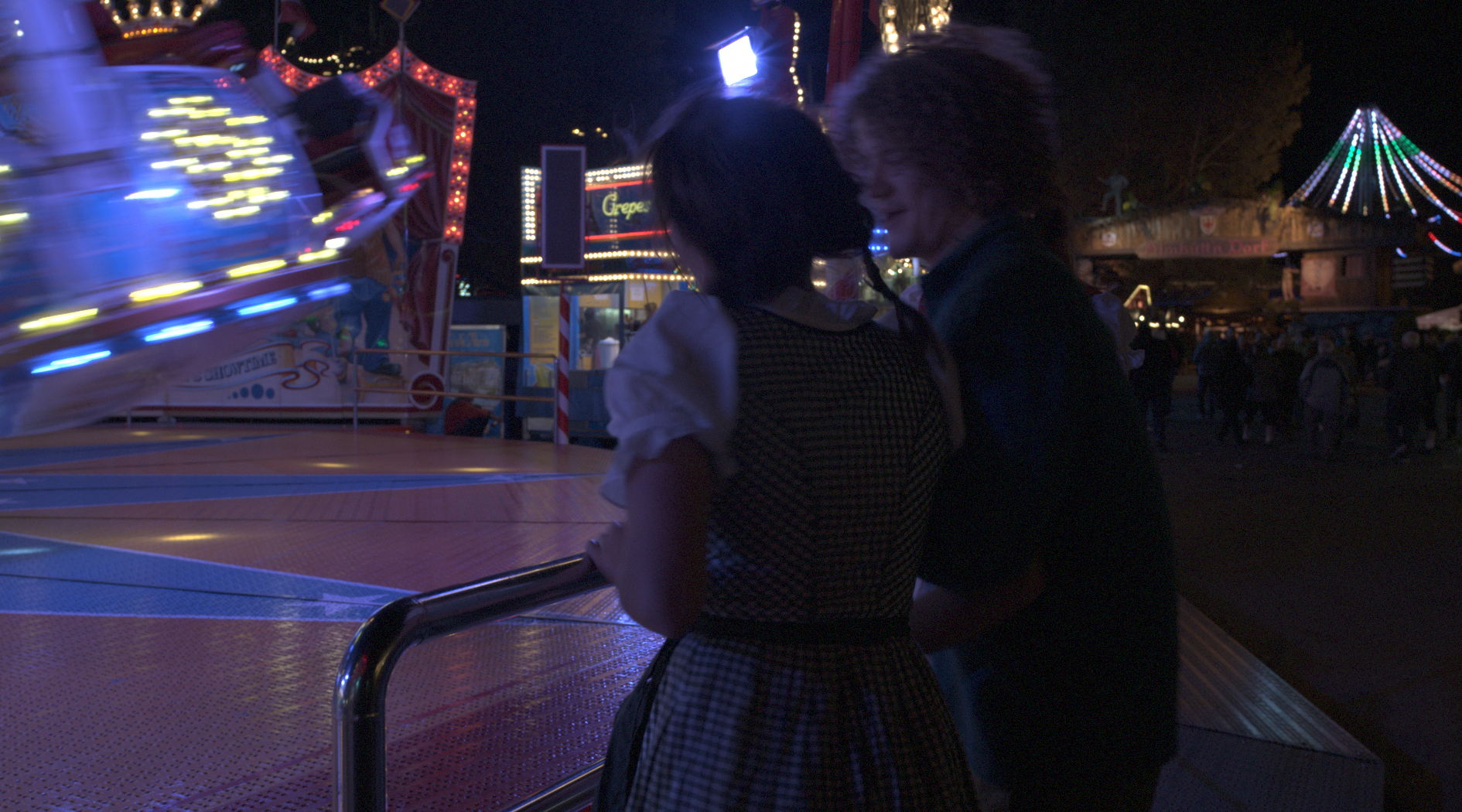}
\includegraphics[trim=600pt 100 400 100,clip, width=2.8cm]{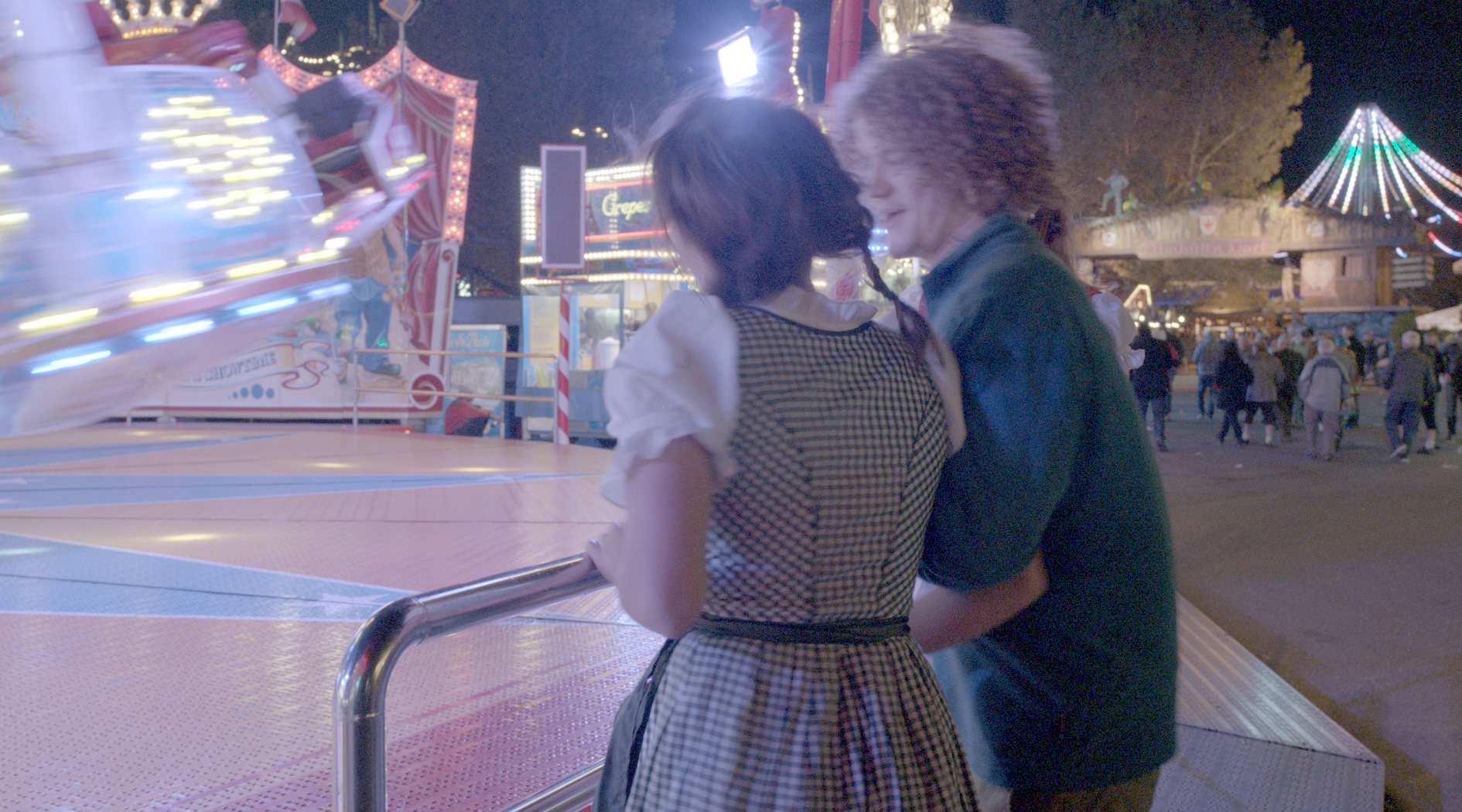}

\caption{Extract of a Stuttgart frame with short exposure and its corresponding ground truth. The ground truth contains noisy areas which makes the PSNR not reliable in this case.}\label{fig:3}
\end{figure}

\subsection{Experimental setup}

\textbf{Datasets.} Following prior works we used Vimeo-90K \cite{xue2019video} dataset in order to generate our training data. We generate alternating-exposure sequences using a simulation pipeline following the same procedure used by most recent papers in the field \cite{chung2023lan,xu2024hdrflow, yue2026f2hdr,  chen2021hdr}. During the training process, we select subsets of three consecutive frames. Compared to other approaches, we select only consecutive frames since we are explicitly interested in video HDR. 
We extract $128\times 128$ patches of the three images at the same spatial position.  A noise of standard deviation {$\sigma=0.001$} is added to the the radiances, which belong to   $[0 , 1]$.

\medskip

To evaluate the proposed method and comparing with existing literature we use the test sequences of Vimeo-90K \cite{xue2019video} and Cinematic Video datasets \cite{froehlich2014creating}. This second dataset consists on HDR video sequences from which we select three consecutive frames and simulate the corresponding LDR alternating the exposure. 

\medskip

\noindent \textbf{Implementation details. }
We implement our framework in PyTorch and train the network end-to-end on a NVIDIA L40 GPU. We employ the AdamW optimizer with exponential decay rates $\beta_1 = 0.9$ and $\beta_2 = 0.999$. The network is trained for {200} epochs with a batch size of 48. The learning rate is initialized at $1 \times 10^{-4}$ and regulated using a cosine annealing scheduler 
to progressively decay it. The $\mu$ parameter of the differentiable $\mu$-law tone mapping function \eqref{tonemap_mu} is set to $5000$, and the gamma correction coefficient $\gamma$ is set to $2.2$ to map alternating LDR inputs into the linear HDR domain. 

\subsection{Comparsion with state-of-the-art}
We compare the proposed method with state-of-the-art methods. In particular, we compare with Deep HDR  \cite{chen2021hdr},  F2HDR \cite{yue2026f2hdr},  HDR Flow  \cite{xu2024hdrflow}, and  LAN HDR \cite{chung2023lan}. We take their corresponding implementations and evaluate all the methods on Vimeo and Stuttgart sequences. 

{Table \ref{table1} presents the PSNR metric for all evaluated methods, with results reported separately for short and long exposures averaged across all sequences. On the Vimeo dataset, our method achieves state-of-the-art performance across all metrics. On the Stuttgart dataset, our approach yields the highest PSNR for long exposures and the second best for short exposures. It is worth noting that LAN-HDR and Deep-HDR both leverage five consecutive frames as input, and Deep-HDR was explicitly trained on Stuttgart data, giving it a distinct advantage in short Stuttgart exposures which have severe noise. We must also observe that the ground truth for Stuttgart short exposures contains noisy areas (see Figure \ref{fig:3}), which makes PSNR values less reliable for such cases.
} 

\begin{table}[ht]
\centering
\scriptsize
\begin{tabular}{llrrrrr}
\toprule
Dataset & Metric & \makecell{Chen\cite{chen2021hdr}} & \makecell{F2HDR\\ \cite{yue2026f2hdr}} & \makecell{HDR\\Flow\cite{xu2024hdrflow}} & \makecell{LAN\\HDR \cite{chung2023lan}} & \makecell{Ours} \\
\midrule
\multirow{3}{*}{Vimeo noise} &
 All & 43.55 & \underline{45.64} & 42.79 & 42.97 & \textbf{48.98} \\
 & High & 39.38 & \underline{39.92} & 38.37 & 38.08 &\textbf{44.83} \\
& Low & 47.72 & \underline{51.36} & 47.21 & 47.86 & \textbf{53.12} \\
\midrule
\multirow{3}{*}{Stuttgart noise} &
 All & 41.06 & 40.24 & 40.85 & \underline{41.37} & \textbf{43.29} \\
 & High & 46.65 & 44.60 & 46.36 & \underline{47.12} &\textbf{50.77} \\
& Low & 35.47 & \textbf{35.88} & 35.34 & 35.62 & \underline{35.81} \\
\bottomrule
\end{tabular}
\caption{Average PSNR values for low-exposure, high-exposure, and all frames across each dataset. 
} \label{table1}
\end{table}

Figures \ref{fig:1} and \ref{fig:2} compare the different methods visually on the Vimeo and Stuttgart databases. We display the processed images as well as the error with respect to the ground truth. The use of the mid-flow permits correct alignment even when areas on the reference frame are saturated to white, as in the woman's face in the Vimeo example and the carousel lights in the Stuttgart one.

\section{Conclusion}
\label{sec:conclusions}

We presented a novel approach to HDR video  from alternating-exposure sequences, addressing the fundamental challenge of accurate inter-frame alignment under varying exposure conditions. Our dual-stream registration strategy combines direct optical flow estimation on equalized LDR frames with a midpoint displacement complement derived from flow computed between the two neighboring frames. The five resulting images — central and neighboring frames with their respective registrations — are then fused through a pyramid-based scheme to produce the final HDR output.

Experimental evaluation confirms that dual-stream registration allows the method to adaptively leverage the most reliable alignment information available at each pixel, yielding substantially improved performance over state-of-the-art HDR reconstruction methods.

Future work may explore extending this framework to a larger number of different exposures, learning-based flow estimation tailored to video HDR, the use of a realistic noise model and large displacement.




\bibliographystyle{IEEEbib}
\bibliography{references}

\end{document}